\documentclass[sigconf,nonacm]{acmart}

\usepackage{balance}

\AtBeginDocument{%
  \providecommand\BibTeX{{%
    \normalfont B\kern-0.5em{\scshape i\kern-0.25em b}\kern-0.8em\TeX}}}

\copyrightyear{2023} 
\acmYear{2023} 
\setcopyright{acmlicensed}\acmConference[AutomotiveUI '23]{15th International Conference on Automotive User Interfaces and Interactive Vehicular Applications}{September 18--22, 2023}{Ingolstadt, Germany}
\acmBooktitle{15th International Conference on Automotive User Interfaces and Interactive Vehicular Applications (AutomotiveUI '23), September 18--22, 2023, Ingolstadt, Germany}
\acmPrice{15.00}
\acmDOI{10.1145/3580585.3607153}
\acmISBN{979-8-4007-0105-4/23/09}

\begin{document}

\title[Personalized in-Vehicle Gesture Recognition with a Time-of-Flight Camera]{It's all about you: Personalized in-Vehicle Gesture Recognition with a Time-of-Flight Camera}


\author{Amr Gomaa}
\orcid{0000-0003-0955-3181}
\authornotemark[1] 
\affiliation{%
  \institution{German Research Center for Artificial Intelligence (DFKI)}
    \city{Saarbr{\"u}cken}
  \country{Germany}
}
\affiliation{%
  \institution{Saarland Informatics Campus}
  \city{Saarbr{\"u}cken}
  \country{Germany}
}
\email{amr.gomaa@dfki.de}

\author{Guillermo Reyes}
\orcid{0000-0003-4064-8605}
\authornote{Both authors contributed equally to this research.}
\affiliation{%
  \institution{German Research Center for Artificial Intelligence (DFKI)}
  \city{Saarbr{\"u}cken}
  \country{Germany}
}
\email{guillermo.reyes@dfki.de}

\author{Michael Feld}
\orcid{0000-0001-6755-5287}
\affiliation{%
  \institution{German Research Center for Artificial Intelligence (DFKI)}
  \city{Saarbr{\"u}cken}
  \country{Germany}
}
\email{michael.feld@dfki.de}

\renewcommand{\shortauthors}{Gomaa et al.}

\begin{abstract}

Despite significant advances in gesture recognition technology, recognizing gestures in a driving environment remains challenging due to limited and costly data and its dynamic, ever-changing nature. In this work, we propose a model-adaptation approach to personalize the training of a CNNLSTM model and improve recognition accuracy while reducing data requirements. Our approach contributes to the field of dynamic hand gesture recognition while driving by providing a more efficient and accurate method that can be customized for individual users, ultimately enhancing the safety and convenience of in-vehicle interactions, as well as driver's experience and system trust. We incorporate hardware enhancement using a time-of-flight camera and algorithmic enhancement through data augmentation, personalized adaptation, and incremental learning techniques. We evaluate the performance of our approach in terms of recognition accuracy, achieving up to 90\%, and show the effectiveness of personalized adaptation and incremental learning for a user-centered design.

\end{abstract}



\maketitle

\section{Introduction}

The last decade has brought significant breakthroughs in speech recognition, image recognition, semantic segmentation, and many other domains. These breakthroughs have been significantly due to the advancement in deep learning (DL) techniques. In particular, this is due to more powerful computing hardware and larger datasets~\cite{lecun2015deep}, but also new ideas and architectures~\cite{szegedy2015going}. 
Gesture recognition is one of these domains. Gestures are a natural form of human interaction, but teaching machines to recognize gestures (particularly dynamic ones) can be challenging despite these advances.
Some of these challenges relate to the technology used to record the data. Although RGB cameras are commonly used in gesture recognition, they are not optimal for the dynamic situation of in-vehicle interaction. That is because they cannot handle poor lighting conditions (e.g., at night) and hands' high-speed motion, which require capturing devices with high frame rates. 

Alternatively, time-of-flight (ToF) depth cameras do not suffer from these problems and can be used with a high frame rate in day and night conditions. On the other hand, ToF cameras can produce noisy results, where the depth of individual pixels changes slightly across frames, or some information is lost. 
Another problem is that DL architectures are known to require large amounts of data. This is primarily due to the high number of parameters required to train an accurate model. The data collection process and resources needed to train a DL model can be extremely costly. Some techniques, like transfer learning, can help reduce the data needed to train a DL model by adapting a pre-trained model from a similar task to a new one. However, its effectiveness depends on how similar the data from both tasks are. In this paper, we are considering the specific use cases of in-vehicle Human Machine Interaction(HMI) gesture interaction using a ToF camera. The first challenge is the lack of datasets for the specific camera view in an in-vehicle environment for hand gesture recognition. Another challenge is that, as far as we know, in-vehicle hand gesture recognition datasets utilize an RGB camera instead of a ToF one.
Since there are essential differences between RGB and depth data, traditional transfer learning is not directly feasible. 

Furthermore, even in the best-case scenario in which it is possible to train a DL dynamic gesture recognition model using ToF cameras with little data, there will always be individual differences in how users perform the gestures. This, in turn, causes the gestures of some individuals not to be as well recognized as those of others. For this reason, gathering data pertaining to each user's individual differences to train personalized models is essential. We gather inspiration for personalization from recent studies on the topic of human-centered artificial intelligence (HCAI), which is gaining rapid and significant interest among researchers in both artificial intelligence (AI) and human-computer interaction (HCI)~\cite{xu2019toward,nowak2018assessing,bryson2019society,shneiderman2020human}. 
Finally, this paper tackles the previously mentioned challenges using a Convolutional Long-Short Term Memory Neural Network (CNNLSTM)~\cite{tsironi2016gesture} and several user-specific adaptations and incremental learning techniques. In particular, we first collect a new dynamic hand gesture data set inside a car using a ToF camera.
Then, we highlight the effect of data augmentation techniques on enhancing the model's accuracy and present different incremental learning adaptation strategies for user-centered gesture recognition. 
Thus, \textbf{this paper's contribution has several folds, as follows}. 1)~We study the feasibility of hand gesture recognition for in-vehicle interaction as conceptualized by modern car manufacturers using a ToF camera instead of RGB; 2)~we propose several essential reproducible preprocessing techniques for ToF cameras as a guideline for future use that is applicable for the automotive domain specifically and other domains generally; 3)~we exploit individual differences in gesture performance by utilizing several personalization and model adaptation techniques such as transfer learning, data augmentation, and incremental learning to enhance the recognition accuracy.

\section{Related Work}

The first form of communication we learn as infants is hand gestures, even before learning to speak, which is why gestures are the most natural form of communication used by humans~\cite{takemura2018neural}. Furthermore, there are around seven thousand languages spoken in the world~\cite{anderson2004many}. However, simple hand gestures (e.g., mid-air gestures, pointing, waving, etc.) are a common form of communication among people, making it more understandable for machines.
Therefore, researchers have tried to incorporate hand gestures into various domains~\cite{Zhao2019implGestureInfotain,Ye2019,Haria2017,Singh2015,Rempel2014,nickel20043d,jing2013human,kehl2004real,Mallika2023,Latif2023,DANG2022improvedgesture,Phyo2019HRIGestures}. Specifically for the automotive domain, researchers attempted to control the infotainment and various parts inside the vehicle~\cite{molchanov2015hand, molchanov2015multi, ohn2014hand, zobl2003real,pickering2007research,Ahmad2018a,Fariman2016,Roider2017,feld2016combine,roider2018see,Stiegemeier2022,Ch2022} as well as interact with objects outside the vehicle~\cite{rumelin2013free,fujimura2013driver,Moniri2012a,Gomaa2020,Aftab2020,gomaa2021ml,gomaa2022adaptive}. They chose hand gestures for several reasons such as: the simplicity and naturalness of hand gestures when interacting with a somewhat complicated machine like a modern car; and the reduced cognitive load on the user when using hand gestures while driving a vehicle (which should be the main focus of the driver).

Several researchers have investigated hand tracking and gesture recognition for more than 30 years. Zimmerman et al.~\cite{Zimmerman1986hand} created a hand glove augmented with analog flux sensors to measure finger bending and track simple gesture interactions. Takahashi and Kishino~\cite{Takahashi1991hand} conducted several studies using this glove-based device to determine the most commonly used gestures by users. Subsequently, several researchers conducted similar studies for glove-based gesture recognition, as highlighted by Dipietro et al.~\cite{dipietro2008survey} in their survey. However, these approaches suffer from several problems, such as the need to wear intrusive gadgets, sensor failures, and inaccuracies. Alternatively, researchers attempted vision-based approaches for hand gesture recognition. Min et al.~\cite{min1997hand}, Rigoll et al.~\cite{rigoll1997high}, Eickeler et al.~\cite{eickeler1998hidden}, and Chen et al.~\cite{chen2003hand} utilized Hidden Markov Models (HMMs) for training and accurate classification of the desired hand symbol (i.e., gesture) under stationary background and fixed light conditions. Similarly, Ren et al.~\cite{ren2009hand}, and Huang et al.~\cite{huang2011gabor} attempted the same task by utilizing image processing techniques (e.g., Gabor filter) and machine learning approaches (e.g., support vector machine (SVM)) to segregate and classify different hand gestures. However, while these approaches supported real-time recognition, they mainly focused on simple static hand gestures with a clear differentiation among them. 

More recently, several vision-based approaches for hand gesture recognition have emerged due to the massive breakthrough in image processing, computer vision, and object recognition using deep neural networks. Stergiopoulou et al.~\cite{stergiopoulou2009hand} utilized a Feed Forward Neural Network (FFNN) for right-hand gesture recognition using a shape-fitting technique. Similarly, Mang~\cite{maung2009real} utilized an FFNN to classify hand gestures after applying a preprocessing feature extraction method (e.g., Oriented Histograms) on hand images with no background information (i.e., cutout hand images with black background). Alternatively, Nagi et al.~\cite{nagi2011max}, Lin et al.~\cite{lin2014human}, Li et al.~\cite{li2019hand}, and Pinto et al.~\cite{pinto2019static} employed a Convolution Neural Network (CNN) learning approach for the recognition due to its high performance on images for several computer vision applications. However, these approaches focus on static gesture recognition scenarios and suffer from limited background information and heavy irreproducible pre-processing techniques. Thus, Maraqa and Abu-Zaiter~\cite{maraqa2008recognition}, Neverova et al.~\cite{neverova2013multi}, K{\"o}p{\"u}kl{\"u} et al.~\cite{kopuklu2019real}, and Molchanov et al.~\cite{molchanov2015hand} investigated Recurrent Neural Network (RNN) and 3D CNN (instead of the traditional 2D CNN approach) to include time series analysis and dynamic gesture recognition, while Tsironi et al.~\cite{tsironi2016gesture} explored the combination of a CNN and RNN into a new architecture named CNNLSTM (an approach first used by Sainath et al.~\cite{sainath2015convolutional} for voice recognition). In this work, we employ a similar CNNLSTM architecture as the state-of-the-art approach for dynamic gesture recognition and introduce easily reproducible preprocessing techniques for ToF cameras.

Finally, while previous approaches have addressed many dynamic gesture recognition challenges, they were still lacking in terms of recognition accuracy. We attribute this to two main problems: \textit{Hardware} and \textit{User-specific Variations}. For the hardware part, most of the previous work either used an RGB camera or a low-performance depth camera (i.e., Kinect). In our approach, we utilized a prototype high-resolution 3D time-of-flight camera (i.e., a high-fidelity depth camera) that is already utilized in high-end modern vehicles~\cite{Aftab2020,2019BMW2019,2021Mercedes-Benz2021}. As for the user-specific variations, most of the current work relies on a one-model-fits-all approach where no adaptation or user-specific personalization are made to the learning model. In contrast, we take inspiration from the Human-centered artificial intelligence (HCAI) domain~\cite{xu2019toward,nowak2018assessing,bryson2019society,shneiderman2020human}, the incremental learning domain~\cite{gepperth2016incremental}, and the transfer learning techniques to employ a user-specific personalized approach that is adaptable for drivers' individual behavior when performing dynamic hand gestures.

\begin{figure*}[t]
	\begin{center}
		\includegraphics[width=0.9\linewidth]{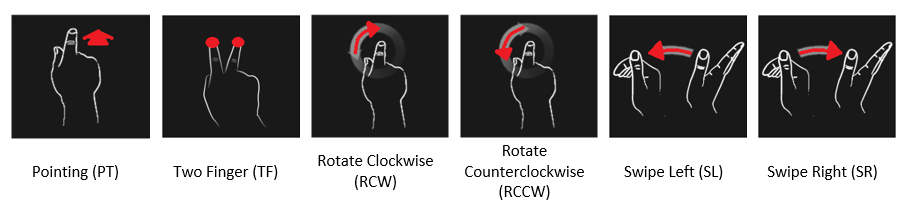}
	\end{center}
	\caption{The six different gestures performed in our dataset and utilized in the the adapted models and incremental learning.}
	\Description{The image shows six sub-images of hands performing a gesture. From left to right, they are \textit{pointing}, \textit{two-fingers}, \textit{rotate clockwise}, \textit{rotate counterclockwise}, \textit{swipe left}, and \textit{swipe right}}
	\label{fig:gestures}
\end{figure*}

\section{Design and Procedure}

In this section, the data acquisition process is highlighted, starting with the environment setup, participants' demographics, and data collection procedure and ending with the detailed description of the final collected dataset, including the split for both the general and personalized models.

\subsection{Apparatus}

The first challenge of in-vehicle hand gesture recognition data acquisition is the necessary instrumentation and technology. As previously mentioned, existing wearable hand-tracking devices are not practical in highly dynamic driving situations, as well as RGB cameras. On the other hand, ToF-based in-vehicle datasets are not available. Thus, we collect our own data set for this purpose. We utilize a state-of-the-art non-commercial hand-tracking camera prototype specially designed for in-vehicle control. This ToF camera was attached to the ceiling of a real car from the inside. It was positioned near the front center, above the gear shift, with a top-down perspective without blocking the view of the windshield. This position corresponds to the position of the gesture camera of recent modern cars such as the \textit{BMW Natural User Interface}~\cite{2019BMW2019} and \textit{Mercedes-Benz User Experience (MBUX) Travel Knowledge Feature}~\cite{2021Mercedes-Benz2021}, which provides a realistic and practical setting applicable to existing in-vehicle environments.

\subsection{Participants}

We recruited 83 participants ($41 \%$ female) with a mean age of $25.97$ years ($SD = 5.97$) for the data collection phase. Regarding handedness, $90.4 \%$ of participants were right-handed. Regarding the driving experience of the participants, the participants had their driver's license on average for $6.54$ years ($SD = 5.76$). Additionally, we collected participants' height and hand length for their possible effect on hand gesture recognition. The participants had an average height of $176.84$ centimeter ($SD = 10.79$) and an average of 66 ($SD = 8.6$), 95.26 ($SD = 9.62$), 105.02 ($SD = 10.08$), 97.72 ($SD = 10.39$), and 78.29 ($SD = 8.43$) millimeters for the thumb, forefinger, middle finger, ring finger and pinkie, respectively.

\begin{table}[b]
\centering
\caption{Participant split according to two disjunct groups. The data of participants in the UBM group are used to train the unviresal background model, subsequently adapted to the participants in the adaptation (AS) group. }
\Description{Participant split according to two disjunct groups. The data of participants in the UBM group are used to train the unviresal background model, subsecuntly adapted to the participants in the adaptation group. The baseline dataset contains 51 participants in the UBM set, and 21 in the adaptive set. In the extended dataset, 72 participants are in the UBM set, compared to 11 in the adaptive set.}
\begin{tabular}{lccc}
\hline
Description & UBM & AS  & Total  \\ \hline
Baseline & $51$  & $21$ & $72$  \\ 
Extended & $72$  & $11$ & $83$  \\ \hline
\end{tabular}
\label{tab:participants}
\end{table}

\subsection{Procedure}

Participants were asked to perform a series of gestures repeatedly. Each gesture was performed ten times per driver. Gestures were performed randomly to avoid any confounding factors or learning effects. Participants were instructed to keep their hands on the wheel as if they were driving to emulate real driving scenarios. The participants would then move their hands from the steering wheel to the area directly below the camera, perform the gesture, and move the hand back to the steering wheel. An experimenter sat in the back of the vehicle throughout the experiment to start and stop the camera without interruption, influence, or comments on the performance of participants' gestures.
Although participants were shown pictures representing the gestures at the beginning, they were not instructed exactly how to perform the gesture to avoid influencing them. This made the recorded data set richer in individual variability and more personalized.

\subsection{Dataset}

The final data set consists of six different simple gesture classes. They are \textit{pointing}, \textit{two fingers}, \textit{rotate clockwise}, \textit{rotate counterclockwise}, \textit{swipe left}, and \textit{swipe right}  gestures as seen in~\autoref{fig:gestures}. Additionally, variations to these simple gestures are introduced, such as multiple pointing, swipe, and two-finger gestures after each other and multiple rotations or fractions of rotations in the same direction for the rotate gestures. For all experiments in this work, only simple gestures were used. These simple gestures could be enough to be used for basic infotainment interaction such as raising the volume, skipping songs, etc. However, new gestures could be added using the same methods discussed here for expanded interaction possibilities.
This data set was split into baseline and extended data sets. The extended dataset included additional gestures per participant (approximately 50 per gesture class) for eleven participants that could be contacted to test the model's incremental training and adaptation on these participants. 
Participants were split into two disjunct groups: a Universal Background Model (\textbf{UBM}) group and an Adaptation Set (\textbf{AS}) group as in~\autoref{tab:participants}.

\begin{table*}[t]
\centering
\renewcommand{\arraystretch}{1.5}
\caption{Data split in training validation and evaluation sets in the baseline and extended datasets. }
\Description{Data split in training validation and evaluation sets in the baseline and extended datasets. In the baseline data set, the UBM set has a total of 3065 gestures and the adaptation set 1260. In the extended dataset, the UBM set has a total of 1326 gestures, while the adaptation set has 2652 gestures.}
\begin{tabular}{lcccccccc}
\hline
Dataset & Training & Validation  & $\Sigma$ UBM & Adaptation & Adapt Val & Evaluation & $\Sigma$ AS & $\Sigma$ \\ \hline
 Baseline & $2455$  & $610$ & $3065$ & $630$ & $252$ & $378$ & $1260$ & $4325$  \\ 
Extended & $3710$  & $926$ & $4636$ & $1326$ &  $534$ & $792$ & $2652$ & $7288$ \\ \hline
\end{tabular}
\label{tab:num_gestures}
\end{table*}

Depending on the experiment, participants data were further split within each of these groups into training, validation, and test sets. For a person in the UBM group, the data was split into 80\% training and 20\% validation. On the contrary, for a person in the AS group, the data were split into 50\% further training for adaptation, 20\% validation and 30\% evaluation. In more detail,~\autoref{tab:num_gestures} highlights the number of samples in each split of the data set.

\section{Methodology and Results}
In this section, we describe the experiments performed that improve the recognition rate of the models from random chance to over 90\% accuracy. However, first, we describe the pre-processing steps that apply to all experiments. Then, we describe the CNNLSTM architecture used in these experiments. We then describe the training of a baseline model and show how this model can be adapted to specific users. We then dive into additional techniques that further enhance the model's performance before experimenting with the amount of data needed to effectively personalize the trained models and how to do this incrementally.

\subsection{Preprocessing}
Deep learning architectures are famous for being end-to-end and do not require hand-crafted features. However, typical deep-learning algorithms also require large amounts of data, from thousands to millions of samples. In the lack of data, it is necessary to simplify the problem of deep learning architecture by performing some simple preprocessing steps. The following describes the preprocessing steps we performed. 

\subsubsection{Number of Frames}

Since the data collected contained variations in the sequence length and sampling rate due to inconsistencies in the network at the recording time, the first step was to standardize the number of frames in each training sequence and the frame rate used to train the model. A frame rate of 12 fps and a maximum sequence length of 70 frames were selected because (a) this reduces the number of parameters that need to be trained, decreasing the amount of data needed to train such parameters, and (b) most sequences had a frame rate above 12 fps and a sequence length below 70 frames. Since neural networks expect a fixed size input, gesture sequences that exceeded the maximum sequence length were truncated, and those that fell below the maximum sequence length were zero-padded. Similarly, gesture sequences that exceeded 12 fps were subsampled to this frame rate. Although it may seem that by altering the dataset in this way, some important information might be lost, making the training of the model more difficult, in reality, most of the dropped information does not contribute much to the classification of the gesture, as contiguous frames are highly correlated. Furthermore, most gestures only take one to two seconds to perform. Removing frames that exceed the max sequence length could remove some noise from the dataset, making the resulting data easier to learn.

\subsubsection{Image Processing} 
Aside from fixing the sequence length and subsampling the gesture sequences to 12 fps, a 3D region of interest was defined between $12$ and $500$ centimeters in depth. Values outside this range were considered to be errors or irrelevant. The images were then standardized, and a Gaussian blur was applied. Finally, the images were scaled down from 320x240 to 80x60 pixels, reducing the number of features needed by the model while maintaining the distinguishing visual features of the hand and fingers.

\subsubsection{Hand Segmentation}

An essential final pre-processing step was that of hand segmentation. This allowed the model to finally start learning the hand gestures, as it removed a significant part of the remaining noise in the image. This was done using a simple technique called background subtraction. This consisted of two steps. In the first step, a background image was calculated for each gesture using the first six frames of the frame sequence. The average background image was subtracted from all the following frames. By doing this, pixels where there is no movement are reduced to near-zero values, while those that contain movement (i.e., where the hand gesture is performed) keep relatively large values. After a qualitative visual comparison, this approach showed a more accurate segmented hand compared to existing RGB-based hand segmentation approaches such as the two-frame subtraction technique from Rigol et al.~\cite{rigoll1997high} and the three-frame differencing technique from Tsironi et al.~\cite{tsironi2016gesture}.

\subsection{Architecture}

\begin{figure*}[t]
	\begin{center}
		\includegraphics[width=0.85\linewidth]{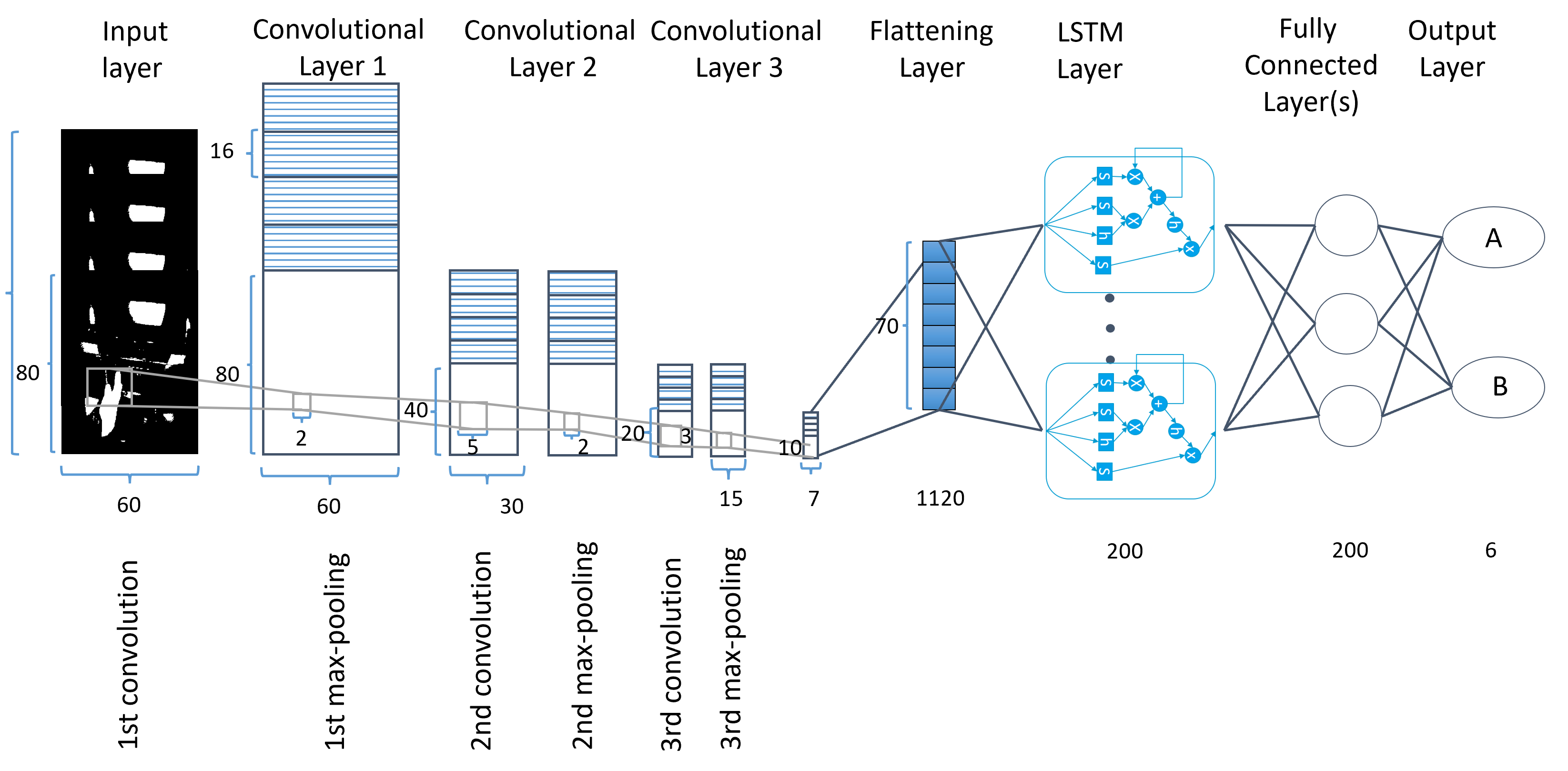}
	\end{center}
	\caption{CNNLSTM architecture}
	\Description{CNNLSTM architecture. The architecture consists of three convolutional layers, followed by an LSTM layer with 200 units, which is itself followed by a fully connected layer, ending with a layer of 6 neuronal layers.}
 \label{fig:architecture}
\end{figure*}

In order to train a model capable of learning dynamic gestures from a ToF camera, we adopted a CNNLSTM architecture, as has been done in similar \cite{tsironi2016gesture} problems. The idea behind CNNLSTM architectures is as follows. A series of convolutional layers extract visual features from the images.  Then, an LSTM layer extracts the time dependencies in these features. Finally, a series of fully connected layers are in charge of classifying the output of the LSTM layer into one of the six gesture types with a softmax activation function. Each convolutional layer consists of a convolution layer, a drop-out layer for regularization, and a max pooling layer. The output of the LSTM layer also goes through a dropout layer for regularization. 
The network was trained with a Tesla P100-SXM2-16GB GPU on simple gesture samples with a batch size of 48. The filter size for the first two convolutional layers was 5x5 and for the third convolutional layer was 3x3.~\autoref{fig:architecture} contains more details about the architecture. 

\subsection{Baseline Training}
\begin{figure}[b]
\includegraphics[width=0.8\linewidth]{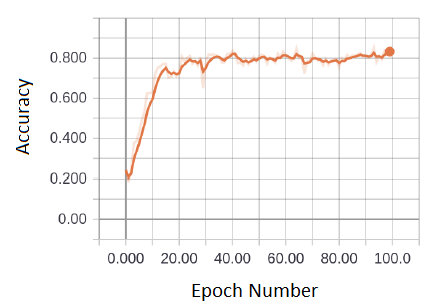}
	\caption{UBM model validation accuracy. The image shows the training progression in epochs. The model has reached its best performance around epoch 30 with a validation accuracy of about 80 percent.}
	\Description{UBM model validation accuracy. The image shows the training progression in epochs. The model has reached its best performance around epoch 30 with a validation accuracy of about 80 percent.}
	\label{fig:ubm_train1}
\end{figure}

Before adapting the models, it was first necessary to create a baseline: the universal background model (UBM). Training the network with the preprocessing steps described above resulted in the model quickly learning to classify the training gestures. \autoref{fig:ubm_train1} shows that the network overfits roughly around epoch 30. However, testing the best model on the baseline evaluation set resulted in an accuracy of 66.9\%. As can be seen in the confusion matrix in \autoref{fig:ubm_as_conf_mtx}, while some gestures, such as the pointing and two-finger gestures, are easy to recognize, differentiating between different directions of the rotate and swipe gestures can be more tricky. This could be due to the lack of data and individual differences in the participants in the universal background model and the participants chosen for the adaptation set. A way to address two of these problems was through adaptation.

\subsection{Single-step Baseline Adaptation}
\begin{figure*}[t]
\begin{center}
		\includegraphics[width=.8\linewidth]{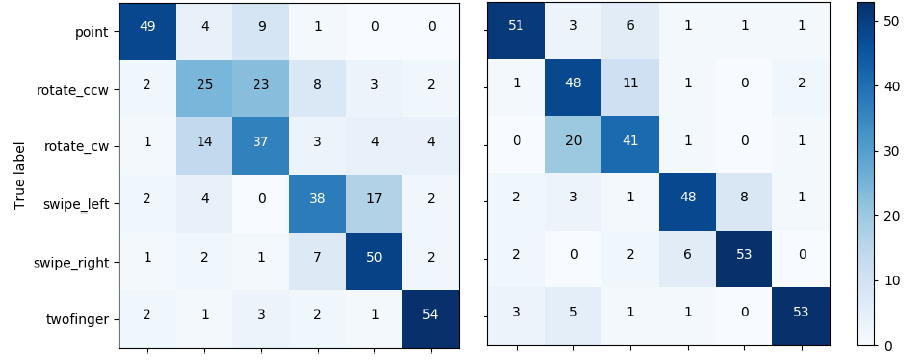}
	\end{center}
	\caption{Left: Confusion matrix on baseline evaluation data for UBM (66.9\% accuracy).  Right: Confusion matrix on baseline evaluation data after single-step adaptation for AS (77.8\% accuracy).}
	\Description{Left: Confusion matrix on baseline evaluation data (66.9\% accuracy).  Right: Confusion matrix on baseline evaluation data after single-step adaptation (77.8\% accuracy). Most of the wrongly classified gestures are found in the rotate and swipe gestures. }
	\label{fig:ubm_as_conf_mtx}
\end{figure*}

While 66.9\% accuracy may not be good enough to interact with a vehicle, we wanted to see if adding some gestures of particular users could improve the recognition rate of those particular users. During the adaptation, the trained UBMs were subsequently trained (or fine-tuned) for the participants in the AS group. The UBM was trained for 20 additional epochs, with a batch size of 12 for each subject individually, using the training and validation data in the adaptive (AS) set. No layers were frozen during this additional training and the model was allowed to learn across all weights. This increased the accuracy of the evaluation data to 77.8\%. As can be seen from the confusion matrix in \autoref{fig:ubm_as_conf_mtx}, the performance improved greatly, especially on the swipe and rotate gestures. However, as this could strongly depend on the subset of participants that make up the UBM and AS sets, we performed cross-validation, changing the participants of each set.  This still resulted in a similar average improvement, with the UBM having 64\% accuracy, compared to 70.7\% after adaptation.

\subsection{Data Augmentation}
While 70.7\% in accuracy is undoubtedly an improvement over 64\%, it is still lacking in human-computer interaction. Since DL models typically require vast amounts of data, a possible way to improve the model is by incorporating additional data into the training. However, getting new data can be expensive and time-consuming. For this reason, finding artificial ways to create more gestures could improve the model's performance. There are different ways to create these artificial data, such as using 3D software to model realistic gestures in a virtual environment, using generative adversarial neural networks, or modifying the collected real dataset itself. The latter is known as data augmentation. In our experiments, we tried a straightforward form of data augmentation consisting of adding random translations along the X and Y axes of -10 to +20 pixels. These simple frame modifications effectively augment the dataset artificially, helping the model's generalization, which resulted in the UBM improving to an average of 79.9\% accuracy and the AS to an average of 86.4\%.

\subsection{Amount of Training Data for Adaptation}
\label{sec:amount_data}
\begin{figure}[b]
\includegraphics[width=0.8\linewidth]{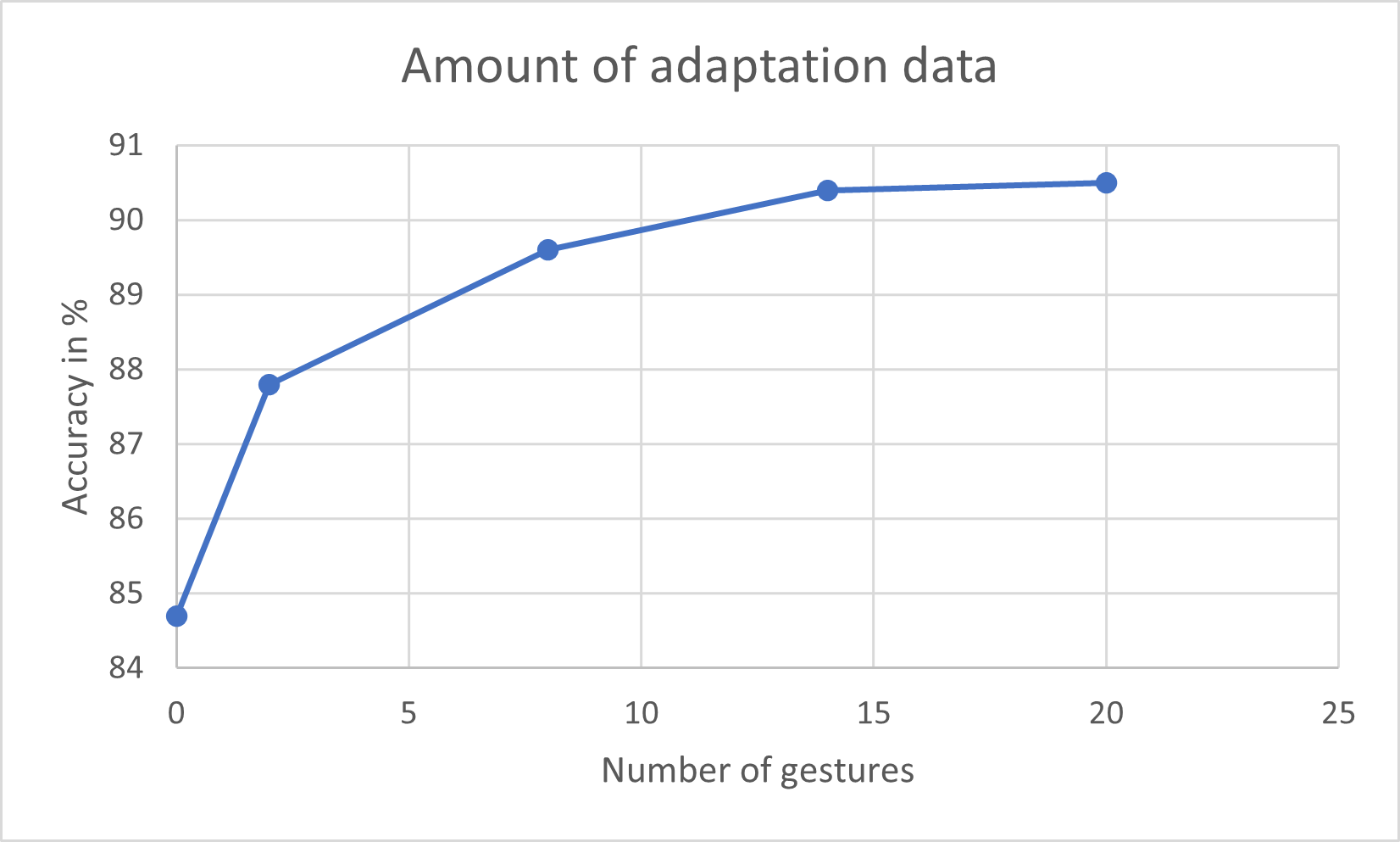}
	\caption{Performance of adapted models according to the number of gestures added to the AS model. The X-axis shows the number of gestures per class with which the model was trained.}
	\Description{Performance of adapted models according to the number of gestures added to the AS model. The X-axis shows the number of gestures per class with which the model was trained. The models were trained with 2, 8, 14, and 20 new gestures per class. These have accuracies of 87.8\%, 89.6\%, 90.4\%, and 90.5\% respectively.}
	\label{fig:amount_data}
\end{figure}

Given that both data augmentation and additional user-specific data benefit the model, an important question is how much data still benefit the model. For these experiments, we trained the models with the extended dataset (see \autoref{tab:num_gestures}). Adding gestures of more participants, together with the data augmentation method, has already increased the accuracy of the UBM to 84.7\%. The next step was to train the AS models with different amounts of new gestures. We trained the AS models using 2, 8, 14, and 20 gestures per gesture class. The results were as expected: the more gestures the AS model was fine-tuned with, the better the performance (see \autoref{fig:amount_data}). The model's performance is already better with only two more gestures per class at 87.8\% accuracy and peaks at around 14 to 20 gestures per class with 90.4\% and 90.5\% accuracy, respectively.

\subsection{Incremental Learning}

\begin{figure*}[t]
\includegraphics[width=\linewidth]{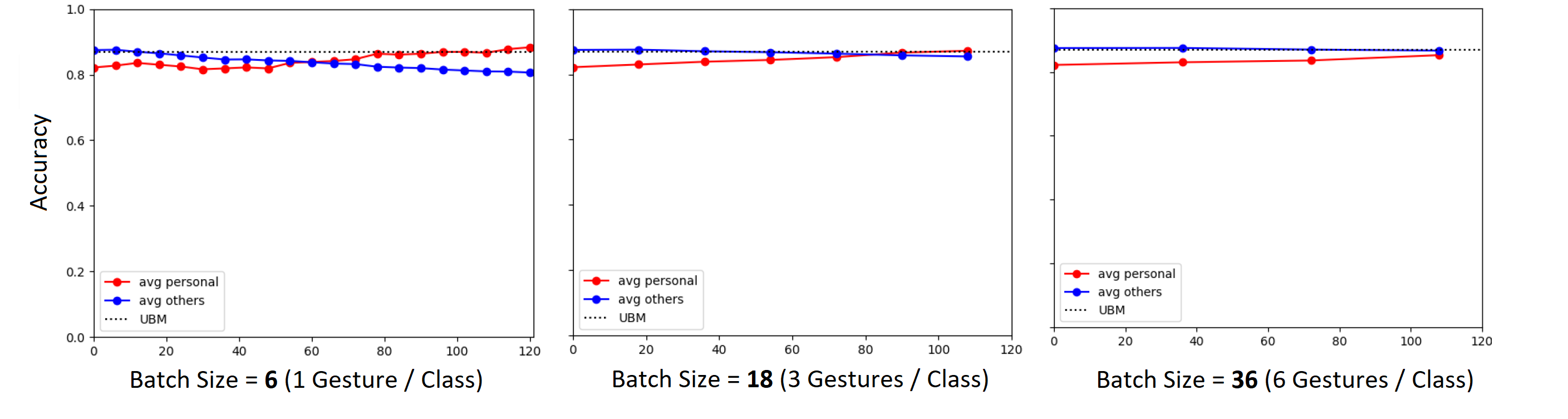}
	\caption{Incremental learning based on different update rates and batch sizes. On the left, one new gesture per class is used for the update. In the middle, three gestures per class. On the right, six gestures per class.}
	\Description{Incremental learning based on different update rates and batch sizes. On the left, one new gesture per class is used for the update. In the middle, three gestures per class. On the right, six gestures per class.}
	\label{fig:inc_batch_size}
\end{figure*}

\begin{figure*}[t]
\includegraphics[width=\linewidth]{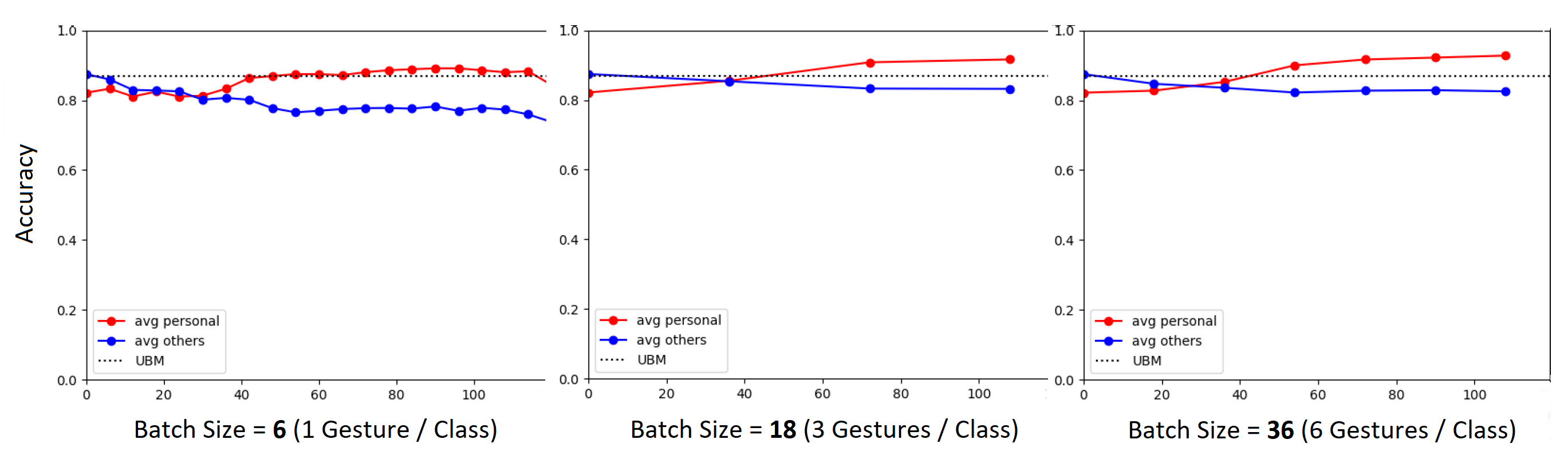}
	\caption{Progression of the incremental learning using validation data. Each data point along the line represents an update step of the network, although the network is updated throughout 15 epochs.}
	\Description{Progression of the incremental learning using validation data. Each data point along the line represents an update step of the network, although the network is updated throughout 15 epochs. The performance achieved seems to be higher for batch sizes 18 and 36. }
	\label{fig:inc_val_data}
\end{figure*}

As seen in \autoref{sec:amount_data}, even small amounts of data can already improve the model. However, it is not always practical to collect the data beforehand. For example, a customer buying a car might be interested in having their personal gestures learned by the system. The customer could then make an appointment and record their gestures, repeating each of the gestures 20 times. They would then need to wait until the system is re-trained and updated with the new gesture recognition model. While this is feasible with a low number of gesture classes, the more tedious this procedure becomes as more gestures are incorporated into the system. Not only that, but every time the manufacturer introduces a new gesture type, the customer would have to return to the agency to repeat the process. A better solution would be if the vehicle itself could learn the user's gestures. This could be done by either the user manually recording and labeling the gestures or by the vehicle automatically labeling the user's gestures during regular interaction. However, how this system could be implemented is beyond the scope of this work. The question we address is how the model could be trained. A problem that might arise from this form of training is that the classifier needs to look at a wide distribution of the data in order to generalize well. If too little data are provided for training, this could have the risk of poor generalization. As one person only uses vehicles at a time, driver-specific models could better detect the gestures of particular drivers. However, if the model is overfitted to specific gestures done by the driver, slight variations in the gestures of the same driver could also be wrongly classified. In what follows, we perform some experiments on how such a model could be trained. In~\autoref{fig:inc_batch_size} and~\autoref{fig:inc_val_data}, the Y-axis represents the accuracy in the percentage of the evaluation set in the extended data set. The \textit{dotted horizontal line} represents the performance of a single UBM from which the adapted models are incrementally trained. The \textit{red line} represents the performance of an adapted model. These models were adapted to a single participant from the 11 participants that make up the adaptive set in the extended data set. The \textit{blue line} shows the performance of the same model tested on the evaluation data of all other participants in the evaluation set (i.e., the remaining ten participants).

\subsubsection{Batch Size}
In the first experiment, three update rates were compared, each corresponding to the batch size for the adaptation. That is, each data point represents an adaptation and also exactly one internal update of the network, using the same number of samples from each gesture class. Furthermore, for all three experiments, the same samples were used in the same order to ensure comparability. The evaluation set consisted of five individuals from the extended set whose results were averaged in this presentation. In particular, persons with less good initial values in the UBM were chosen in order to be able to observe the improvement better. For this reason, the red line in the cases shown also starts below the dashed UBM line. The starting point for adaptation represents the previous adapted network in each case. It was adapted with one single epoch in each batch, and no validation set was used. The lack of validation data was important because, in the real world, the availability of correctly labeled validation data for all persons is not guaranteed. Therefore, we wanted to experiment with the feasibility of adapting the model without validation data.

The following hypotheses were conceptualized for this approach and should be manifest in~\autoref{fig:inc_batch_size}:
\begin{itemize}
    \item H1: The red line should (tend to) increase as the model improves for the adapted person by adding new gestures. 
    \item H2: The blue line should (tend to) decrease because the adapted model now works less well for other persons. (This ``forgetting'' is a general side effect of personalization).
    \item H3: The accuracy at the right end of the X-axis should, in theory, be about the same regardless of the update rates for the red line.
    \item H4: The experiments with small batch size should show more significant fluctuations within the same interval on the X-axis than the graphs with larger batch size (Stability-Plasticity dilemma).
\end{itemize}

As seen in \autoref{fig:inc_batch_size}, the red line is mainly increasing, albeit with fluctuations. The fluctuations are probably mainly due to the fact that no validation set was used, meaning that continuous improvement cannot be guaranteed. This is consistent with hypothesis H1.
Additionally, the blue line decreases as expected (hypothesis H2), although this is mainly the case where a single gesture per class is used. This reflects the stability-plasticity dilemma: a reactive system that adapts quickly to changes in the data also tends to ``forget'' previous data faster. Updates with smaller batch sizes allow for a faster adaptation of the model but also fluctuate more strongly (hypothesis H4)
Although the accuracy achieved at the end of the three settings deviates by up to 3\% from each other, when the batch size 6 setting reaches gesture 108, it closely matches the other settings where they reach the same number. 

It should also be noted that while these results reflect the average improvement, the degree to which the model was adapted for each individual was different. For some participants, the model improved continuously as new data was added. For some, the model performance first went down before going back up; for others, the recognition stayed mainly the same throughout.  Therefore, a significant improvement through adaptation cannot be guaranteed in all cases. Especially with lower update rates, more substantial fluctuations in the model's performance are expected.

\subsubsection{Sample Weight}

In another experiment, we evaluated the influence of the sample weight as a means of correcting for learning speed and, thus, controlling the stability-plasticity of the model. We hypothesized that a lower sample weight should reduce the adaptation rate. This could counterbalance the fast adaptation rate for small batch sizes or the slow adaptation rate for large batch sizes. We compare two settings. In one, the samples are given a weight of $1$, whereas, in the other, the samples are given a weight of $0.5$. 
The results suggest that the higher sample weight indeed results in faster adaptation, and thus it can be used to speed up or slow down the adaptation of the model. 

\subsubsection{Use of Validation Data}
This set of experiments investigates the influence of the update rate when validation data are available and multiple epochs can be performed during adaptation. The models were trained for 15 epochs, and of those, the overall best model is used as the base model for adaptation at each step. A separate set of samples of the adaptation subjects of 8 samples per class is used as the validation set.
It was expected that these models would perform better than the previous models. However, since the best model selected is based only on the validation data, this was not guaranteed. 
With the above-mentioned parameters, deterioration between the data points on the X-axis can be almost completely avoided, i.e., the model continuously improves (see \autoref{fig:inc_val_data}). However, this behavior can only be achieved with a suitable validation set. In practice, however, this would imply that the driver's data would have to be recorded manually by the drivers themselves or in the agency. Additionally, using validation data appears to further improve the performance of the adapted models for the individuals while equally deteriorating the performance for other subjects. This could most likely have to do with the number of epochs for which the models were trained, as each epoch constitutes an update of the model. The decrease in accuracy for other participants is, however, not a problem in the case of intelligent vehicles, as these could store these personalized models for each of the drivers and switch them together with the driving profiles. It is also interesting that in the case of batch sizes 18 and 36, the increase and decrease in performance for the individual and the rest, respectively, seem more stable and pronounced. In the case where the batch size is six, the performance appears worse than when the validation set is not used for both the adapted individual and the rest of the participants. This, again, could have to do with the number of updates (epochs) to the network that are performed at each step.

\section{Conclusion}

This work demonstrates the feasibility of implementing dynamic gesture recognition in the automotive domain, even when data is limited. Although transfer learning can address the issue of limited data, its suitability may be domain-specific due to variations in data properties. To overcome this limitation, we collected a dataset of 4325 depth-based gestures, which was later extended to 7288. However, training a CNNLSTM model required additional data, prompting the development of preprocessing guidelines for time-of-flight sensors to mitigate the data scarcity problem. In addition, we propose data enhancement and incremental learning techniques to adapt the learning model and enhance the accuracy of a universal background model. The significance of these experiments and techniques in the context of in-vehicle human-machine interaction lies in their ability to facilitate model adaptation to the driver's interaction behavior. Furthermore, this study offers valuable insights and guidelines on effectively utilizing limited data to train dynamic gesture recognition algorithms with satisfactory performance, which can be extrapolated to other domains. While the internal validity is maximized in this work, future research will focus on examining the external validity of our model adaptation system in a driving environment, where additional challenges in data acquisition techniques arise.

\begin{acks}

This work is partially funded by the German Ministry of Education and Research (BMBF) under the CAMELOT project (Grant Number: 01IW20008).

\end{acks}

\bibliographystyle{ACM-Reference-Format}
\bibliography{references}


\begin{thebibliography}{62}


\ifx \showCODEN    \undefined \def \showCODEN     #1{\unskip}     \fi
\ifx \showDOI      \undefined \def \showDOI       #1{#1}\fi
\ifx \showISBNx    \undefined \def \showISBNx     #1{\unskip}     \fi
\ifx \showISBNxiii \undefined \def \showISBNxiii  #1{\unskip}     \fi
\ifx \showISSN     \undefined \def \showISSN      #1{\unskip}     \fi
\ifx \showLCCN     \undefined \def \showLCCN      #1{\unskip}     \fi
\ifx \shownote     \undefined \def \shownote      #1{#1}          \fi
\ifx \showarticletitle \undefined \def \showarticletitle #1{#1}   \fi
\ifx \showURL      \undefined \def \showURL       {\relax}        \fi
\providecommand\bibfield[2]{#2}
\providecommand\bibinfo[2]{#2}
\providecommand\natexlab[1]{#1}
\providecommand\showeprint[2][]{arXiv:#2}

\bibitem[\protect\citeauthoryear{??}{201}{2019}]%
        {2019BMW2019}
 \bibinfo{year}{2019}\natexlab{}.
\newblock \bibinfo{title}{{BMW’s Innovative Gesture Control Technology Sets the Industry Standard. Accessed 12-01-2023}}.
\newblock
\newblock
\urldef\tempurl%
\url{https://news.indigoautogroup.com/bmws-innovative-gesture-control-technology-sets-the-industry-standard/}
\showURL{%
\tempurl}


\bibitem[\protect\citeauthoryear{??}{202}{2021}]%
        {2021Mercedes-Benz2021}
 \bibinfo{year}{2021}\natexlab{}.
\newblock \bibinfo{title}{{The new Mercedes-Maybach S-Class up close: MBUX Interior Assist Rear. Accessed 12-01-2023}}.
\newblock
\newblock
\urldef\tempurl%
\url{https://group-media.mercedes-benz.com/marsMediaSite/en/instance/ko/The-new-Mercedes-Maybach-S-Class-up-close-MBUX-Interior-Assist-Rear.xhtml?oid=50185650}
\showURL{%
\tempurl}


\bibitem[\protect\citeauthoryear{Aftab, von~der Beeck, and Feld}{Aftab et~al\mbox{.}}{2020}]%
        {Aftab2020}
\bibfield{author}{\bibinfo{person}{Abdul~Rafey Aftab}, \bibinfo{person}{Michael von~der Beeck}, {and} \bibinfo{person}{Michael Feld}.} \bibinfo{year}{2020}\natexlab{}.
\newblock \showarticletitle{You Have a Point There: Object Selection Inside an Automobile Using Gaze, Head Pose and Finger Pointing}. In \bibinfo{booktitle}{\emph{Proceedings of the 2020 International Conference on Multimodal Interaction}}. \bibinfo{publisher}{Association for Computing Machinery}, \bibinfo{address}{New York, NY, USA}, \bibinfo{pages}{595–603}.
\newblock
\showISBNx{9781450375818}
\urldef\tempurl%
\url{https://doi.org/10.1145/3382507.3418836}
\showURL{%
\tempurl}


\bibitem[\protect\citeauthoryear{Ahmad, Hare, Singh, Shabani, Lindsay, Skrypchuk, Langdon, and Godsill}{Ahmad et~al\mbox{.}}{2018}]%
        {Ahmad2018a}
\bibfield{author}{\bibinfo{person}{Bashar~I. Ahmad}, \bibinfo{person}{Chrisminder Hare}, \bibinfo{person}{Harpreet Singh}, \bibinfo{person}{Arber Shabani}, \bibinfo{person}{Briana Lindsay}, \bibinfo{person}{Lee Skrypchuk}, \bibinfo{person}{Patrick Langdon}, {and} \bibinfo{person}{Simon Godsill}.} \bibinfo{year}{2018}\natexlab{}.
\newblock \showarticletitle{{Selection facilitation schemes for predictive touch with mid-air pointing gestures in automotive displays}}. In \bibinfo{booktitle}{\emph{Proceedings of the 10th International Conference on Automotive User Interfaces and Interactive Vehicular Applications}}. \bibinfo{publisher}{ACM}, \bibinfo{pages}{21--32}.
\newblock
\showISBNx{9781450359467}


\bibitem[\protect\citeauthoryear{Anderson}{Anderson}{2010}]%
        {anderson2004many}
\bibfield{author}{\bibinfo{person}{Stephen~R Anderson}.} \bibinfo{year}{2010}\natexlab{}.
\newblock \showarticletitle{How many languages are there in the world}.
\newblock \bibinfo{journal}{\emph{Linguistic Society of America}} (\bibinfo{year}{2010}).
\newblock
\newblock
\shownote{6 pages.}


\bibitem[\protect\citeauthoryear{Bryson and Theodorou}{Bryson and Theodorou}{2019}]%
        {bryson2019society}
\bibfield{author}{\bibinfo{person}{Joanna~J. Bryson} {and} \bibinfo{person}{Andreas Theodorou}.} \bibinfo{year}{2019}\natexlab{}.
\newblock \bibinfo{booktitle}{\emph{How Society Can Maintain Human-Centric Artificial Intelligence}}.
\newblock \bibinfo{publisher}{Springer Singapore}, \bibinfo{address}{Singapore}, \bibinfo{pages}{305--323}.
\newblock
\showISBNx{978-981-13-7725-9}


\bibitem[\protect\citeauthoryear{Ch, Tosca, Crump, Ansah, Kun, and Shaer}{Ch et~al\mbox{.}}{2022}]%
        {Ch2022}
\bibfield{author}{\bibinfo{person}{Nabil Al~Nahin Ch}, \bibinfo{person}{Diana Tosca}, \bibinfo{person}{Tyanna Crump}, \bibinfo{person}{Alberta Ansah}, \bibinfo{person}{Andrew Kun}, {and} \bibinfo{person}{Orit Shaer}.} \bibinfo{year}{2022}\natexlab{}.
\newblock \showarticletitle{Gesture and Voice Commands to Interact With AR Windshield Display in Automated Vehicle: A Remote Elicitation Study}. In \bibinfo{booktitle}{\emph{Proceedings of the 14th International Conference on Automotive User Interfaces and Interactive Vehicular Applications}} (Seoul, Republic of Korea) \emph{(\bibinfo{series}{AutomotiveUI '22})}. \bibinfo{publisher}{Association for Computing Machinery}, \bibinfo{address}{New York, NY, USA}, \bibinfo{pages}{171–182}.
\newblock
\showISBNx{9781450394154}
\urldef\tempurl%
\url{https://doi.org/10.1145/3543174.3545257}
\showDOI{\tempurl}


\bibitem[\protect\citeauthoryear{Chen, Fu, and Huang}{Chen et~al\mbox{.}}{2003}]%
        {chen2003hand}
\bibfield{author}{\bibinfo{person}{Feng-Sheng Chen}, \bibinfo{person}{Chih-Ming Fu}, {and} \bibinfo{person}{Chung-Lin Huang}.} \bibinfo{year}{2003}\natexlab{}.
\newblock \showarticletitle{Hand gesture recognition using a real-time tracking method and hidden Markov models}.
\newblock \bibinfo{journal}{\emph{Image and vision computing}} \bibinfo{volume}{21}, \bibinfo{number}{8} (\bibinfo{year}{2003}), \bibinfo{pages}{745--758}.
\newblock


\bibitem[\protect\citeauthoryear{Dang, Tran, Nguyen, Kim, and Monet}{Dang et~al\mbox{.}}{2022}]%
        {DANG2022improvedgesture}
\bibfield{author}{\bibinfo{person}{Tuan~Linh Dang}, \bibinfo{person}{Sy~Dat Tran}, \bibinfo{person}{Thuy~Hang Nguyen}, \bibinfo{person}{Suntae Kim}, {and} \bibinfo{person}{Nicolas Monet}.} \bibinfo{year}{2022}\natexlab{}.
\newblock \showarticletitle{An improved hand gesture recognition system using keypoints and hand bounding boxes}.
\newblock \bibinfo{journal}{\emph{Array}} (\bibinfo{year}{2022}), \bibinfo{pages}{100251}.
\newblock
\showISSN{2590-0056}
\urldef\tempurl%
\url{https://doi.org/10.1016/j.array.2022.100251}
\showDOI{\tempurl}


\bibitem[\protect\citeauthoryear{Dipietro, Sabatini, and Dario}{Dipietro et~al\mbox{.}}{2008}]%
        {dipietro2008survey}
\bibfield{author}{\bibinfo{person}{Laura Dipietro}, \bibinfo{person}{Angelo~M Sabatini}, {and} \bibinfo{person}{Paolo Dario}.} \bibinfo{year}{2008}\natexlab{}.
\newblock \showarticletitle{A survey of glove-based systems and their applications}.
\newblock \bibinfo{journal}{\emph{IEEE transactions on systems, man, and cybernetics, part c (applications and reviews)}} \bibinfo{volume}{38}, \bibinfo{number}{4} (\bibinfo{year}{2008}), \bibinfo{pages}{461--482}.
\newblock


\bibitem[\protect\citeauthoryear{Eickeler, Kosmala, and Rigoll}{Eickeler et~al\mbox{.}}{1998}]%
        {eickeler1998hidden}
\bibfield{author}{\bibinfo{person}{Stefan Eickeler}, \bibinfo{person}{Andreas Kosmala}, {and} \bibinfo{person}{Gerhard Rigoll}.} \bibinfo{year}{1998}\natexlab{}.
\newblock \showarticletitle{Hidden markov model based continuous online gesture recognition}. In \bibinfo{booktitle}{\emph{Proceedings. Fourteenth International Conference on Pattern Recognition (Cat. No. 98EX170)}}, Vol.~\bibinfo{volume}{2}. IEEE, \bibinfo{pages}{1206--1208}.
\newblock


\bibitem[\protect\citeauthoryear{Fariman, Alyamani, Kavakli, and Hamey}{Fariman et~al\mbox{.}}{2016}]%
        {Fariman2016}
\bibfield{author}{\bibinfo{person}{Hessam~Jahani Fariman}, \bibinfo{person}{Hasan~J. Alyamani}, \bibinfo{person}{Manolya Kavakli}, {and} \bibinfo{person}{Len Hamey}.} \bibinfo{year}{2016}\natexlab{}.
\newblock \showarticletitle{{Designing a user-defined gesture vocabulary for an in-vehicle climate control system}}. In \bibinfo{booktitle}{\emph{Proceedings of the 28th Australian Computer-Human Interaction Conference}}. \bibinfo{publisher}{ACM}, \bibinfo{pages}{391--395}.
\newblock
\showISBNx{9781450346184}


\bibitem[\protect\citeauthoryear{Fujimura, Xu, Tran, Bhandari, and Ng-Thow-Hing}{Fujimura et~al\mbox{.}}{2013}]%
        {fujimura2013driver}
\bibfield{author}{\bibinfo{person}{Kikuo Fujimura}, \bibinfo{person}{Lijie Xu}, \bibinfo{person}{Cuong Tran}, \bibinfo{person}{Rishabh Bhandari}, {and} \bibinfo{person}{Victor Ng-Thow-Hing}.} \bibinfo{year}{2013}\natexlab{}.
\newblock \showarticletitle{{Driver queries using wheel-constrained finger pointing and 3-D head-up display visual feedback}}. In \bibinfo{booktitle}{\emph{Proceedings of the 5th International Conference on Automotive User Interfaces and Interactive Vehicular Applications}}. \bibinfo{publisher}{ACM}, \bibinfo{pages}{56--62}.
\newblock


\bibitem[\protect\citeauthoryear{Gepperth and Hammer}{Gepperth and Hammer}{2016}]%
        {gepperth2016incremental}
\bibfield{author}{\bibinfo{person}{Alexander Gepperth} {and} \bibinfo{person}{Barbara Hammer}.} \bibinfo{year}{2016}\natexlab{}.
\newblock \showarticletitle{Incremental learning algorithms and applications}. In \bibinfo{booktitle}{\emph{European symposium on artificial neural networks (ESANN)}}.
\newblock


\bibitem[\protect\citeauthoryear{Gomaa}{Gomaa}{2022}]%
        {gomaa2022adaptive}
\bibfield{author}{\bibinfo{person}{Amr Gomaa}.} \bibinfo{year}{2022}\natexlab{}.
\newblock \showarticletitle{Adaptive User-Centered Multimodal Interaction towards Reliable and Trusted Automotive Interfaces}. In \bibinfo{booktitle}{\emph{Proceedings of the 2022 International Conference on Multimodal Interaction}} (Bengaluru, India) \emph{(\bibinfo{series}{ICMI '22})}. \bibinfo{publisher}{Association for Computing Machinery}, \bibinfo{address}{New York, NY, USA}, \bibinfo{pages}{690–695}.
\newblock
\showISBNx{9781450393904}
\urldef\tempurl%
\url{https://doi.org/10.1145/3536221.3557034}
\showDOI{\tempurl}


\bibitem[\protect\citeauthoryear{Gomaa, Reyes, Alles, Rupp, and Feld}{Gomaa et~al\mbox{.}}{2020}]%
        {Gomaa2020}
\bibfield{author}{\bibinfo{person}{Amr Gomaa}, \bibinfo{person}{Guillermo Reyes}, \bibinfo{person}{Alexandra Alles}, \bibinfo{person}{Lydia Rupp}, {and} \bibinfo{person}{Michael Feld}.} \bibinfo{year}{2020}\natexlab{}.
\newblock \showarticletitle{Studying Person-Specific Pointing and Gaze Behavior for Multimodal Referencing of Outside Objects from a Moving Vehicle}. In \bibinfo{booktitle}{\emph{Proceedings of the 2020 International Conference on Multimodal Interaction}}. \bibinfo{publisher}{Association for Computing Machinery}, \bibinfo{address}{New York, NY, USA}, \bibinfo{pages}{501–509}.
\newblock
\showISBNx{9781450375818}
\urldef\tempurl%
\url{https://doi.org/10.1145/3382507.3418817}
\showURL{%
\tempurl}


\bibitem[\protect\citeauthoryear{Gomaa, Reyes, and Feld}{Gomaa et~al\mbox{.}}{2021}]%
        {gomaa2021ml}
\bibfield{author}{\bibinfo{person}{Amr Gomaa}, \bibinfo{person}{Guillermo Reyes}, {and} \bibinfo{person}{Michael Feld}.} \bibinfo{year}{2021}\natexlab{}.
\newblock \showarticletitle{ML-PersRef: A Machine Learning-Based Personalized Multimodal Fusion Approach for Referencing Outside Objects From a Moving Vehicle}. In \bibinfo{booktitle}{\emph{Proceedings of the 2021 International Conference on Multimodal Interaction}}. \bibinfo{publisher}{Association for Computing Machinery}, \bibinfo{address}{New York, NY, USA}, \bibinfo{pages}{318–327}.
\newblock
\showISBNx{9781450384810}
\urldef\tempurl%
\url{https://doi.org/10.1145/3462244.3479910}
\showDOI{\tempurl}


\bibitem[\protect\citeauthoryear{Haria, Subramanian, Asokkumar, Poddar, and Nayak}{Haria et~al\mbox{.}}{2017}]%
        {Haria2017}
\bibfield{author}{\bibinfo{person}{Aashni Haria}, \bibinfo{person}{Archanasri Subramanian}, \bibinfo{person}{Nivedhitha Asokkumar}, \bibinfo{person}{Shristi Poddar}, {and} \bibinfo{person}{Jyothi~S. Nayak}.} \bibinfo{year}{2017}\natexlab{}.
\newblock \showarticletitle{Hand gesture recognition for human computer interaction}.
\newblock \bibinfo{journal}{\emph{Procedia Computer Science}}  \bibinfo{volume}{115} (\bibinfo{year}{2017}), \bibinfo{pages}{367--374}.
\newblock
\showISSN{1877-0509}


\bibitem[\protect\citeauthoryear{Huang, Hu, and Chang}{Huang et~al\mbox{.}}{2011}]%
        {huang2011gabor}
\bibfield{author}{\bibinfo{person}{Deng-Yuan Huang}, \bibinfo{person}{Wu-Chih Hu}, {and} \bibinfo{person}{Sung-Hsiang Chang}.} \bibinfo{year}{2011}\natexlab{}.
\newblock \showarticletitle{Gabor filter-based hand-pose angle estimation for hand gesture recognition under varying illumination}.
\newblock \bibinfo{journal}{\emph{Expert Systems with Applications}} \bibinfo{volume}{38}, \bibinfo{number}{5} (\bibinfo{year}{2011}), \bibinfo{pages}{6031--6042}.
\newblock


\bibitem[\protect\citeauthoryear{Jing and Ye-Peng}{Jing and Ye-Peng}{2013}]%
        {jing2013human}
\bibfield{author}{\bibinfo{person}{Pan Jing} {and} \bibinfo{person}{Guan Ye-Peng}.} \bibinfo{year}{2013}\natexlab{}.
\newblock \showarticletitle{{Human-computer interaction using pointing gesture based on an adaptive virtual touch screen}}.
\newblock \bibinfo{journal}{\emph{International Journal of Signal Processing, Image Processing and Pattern Recognition}} \bibinfo{volume}{6}, \bibinfo{number}{4} (\bibinfo{year}{2013}), \bibinfo{pages}{81--91}.
\newblock


\bibitem[\protect\citeauthoryear{Kehl and Van~Gool}{Kehl and Van~Gool}{2004}]%
        {kehl2004real}
\bibfield{author}{\bibinfo{person}{Roland Kehl} {and} \bibinfo{person}{Luc Van~Gool}.} \bibinfo{year}{2004}\natexlab{}.
\newblock \showarticletitle{{Real-time pointing gesture recognition for an immersive environment}}. In \bibinfo{booktitle}{\emph{Proceedings of the 6th International Conference on Automatic Face and Gesture Recognition}}. \bibinfo{publisher}{IEEE}, \bibinfo{pages}{577--582}.
\newblock


\bibitem[\protect\citeauthoryear{K{\"o}p{\"u}kl{\"u}, Gunduz, Kose, and Rigoll}{K{\"o}p{\"u}kl{\"u} et~al\mbox{.}}{2019}]%
        {kopuklu2019real}
\bibfield{author}{\bibinfo{person}{Okan K{\"o}p{\"u}kl{\"u}}, \bibinfo{person}{Ahmet Gunduz}, \bibinfo{person}{Neslihan Kose}, {and} \bibinfo{person}{Gerhard Rigoll}.} \bibinfo{year}{2019}\natexlab{}.
\newblock \showarticletitle{Real-time hand gesture detection and classification using convolutional neural networks}. In \bibinfo{booktitle}{\emph{2019 14th IEEE International Conference on Automatic Face \& Gesture Recognition (FG 2019)}}. IEEE, \bibinfo{pages}{1--8}.
\newblock


\bibitem[\protect\citeauthoryear{Latif, Buckley, and Secco}{Latif et~al\mbox{.}}{2023}]%
        {Latif2023}
\bibfield{author}{\bibinfo{person}{Bilawal Latif}, \bibinfo{person}{Neil Buckley}, {and} \bibinfo{person}{Emanuele~Lindo Secco}.} \bibinfo{year}{2023}\natexlab{}.
\newblock \showarticletitle{Hand Gesture and Human-Drone Interaction}. In \bibinfo{booktitle}{\emph{Intelligent Systems and Applications}}, \bibfield{editor}{\bibinfo{person}{Kohei Arai}} (Ed.). \bibinfo{publisher}{Springer International Publishing}, \bibinfo{address}{Cham}, \bibinfo{pages}{299--308}.
\newblock
\showISBNx{978-3-031-16075-2}


\bibitem[\protect\citeauthoryear{LeCun, Bengio, and Hinton}{LeCun et~al\mbox{.}}{2015}]%
        {lecun2015deep}
\bibfield{author}{\bibinfo{person}{Yann LeCun}, \bibinfo{person}{Yoshua Bengio}, {and} \bibinfo{person}{Geoffrey Hinton}.} \bibinfo{year}{2015}\natexlab{}.
\newblock \showarticletitle{Deep learning}.
\newblock \bibinfo{journal}{\emph{nature}} \bibinfo{volume}{521}, \bibinfo{number}{7553} (\bibinfo{year}{2015}), \bibinfo{pages}{436--444}.
\newblock


\bibitem[\protect\citeauthoryear{Li, Tang, Sun, Kong, Jiang, Jiang, Tao, Xu, and Liu}{Li et~al\mbox{.}}{2019}]%
        {li2019hand}
\bibfield{author}{\bibinfo{person}{Gongfa Li}, \bibinfo{person}{Heng Tang}, \bibinfo{person}{Ying Sun}, \bibinfo{person}{Jianyi Kong}, \bibinfo{person}{Guozhang Jiang}, \bibinfo{person}{Du Jiang}, \bibinfo{person}{Bo Tao}, \bibinfo{person}{Shuang Xu}, {and} \bibinfo{person}{Honghai Liu}.} \bibinfo{year}{2019}\natexlab{}.
\newblock \showarticletitle{Hand gesture recognition based on convolution neural network}.
\newblock \bibinfo{journal}{\emph{Cluster Computing}} \bibinfo{volume}{22}, \bibinfo{number}{2} (\bibinfo{year}{2019}), \bibinfo{pages}{2719--2729}.
\newblock


\bibitem[\protect\citeauthoryear{Lin, Hsu, and Chen}{Lin et~al\mbox{.}}{2014}]%
        {lin2014human}
\bibfield{author}{\bibinfo{person}{Hsien-I Lin}, \bibinfo{person}{Ming-Hsiang Hsu}, {and} \bibinfo{person}{Wei-Kai Chen}.} \bibinfo{year}{2014}\natexlab{}.
\newblock \showarticletitle{Human hand gesture recognition using a convolution neural network}. In \bibinfo{booktitle}{\emph{2014 IEEE International Conference on Automation Science and Engineering (CASE)}}. IEEE, \bibinfo{pages}{1038--1043}.
\newblock


\bibitem[\protect\citeauthoryear{Mallika, Ghosh, and Pradhan}{Mallika et~al\mbox{.}}{2023}]%
        {Mallika2023}
\bibfield{author}{\bibinfo{person}{Garg Mallika}, \bibinfo{person}{Debashis Ghosh}, {and} \bibinfo{person}{Pyari~Mohan Pradhan}.} \bibinfo{year}{2023}\natexlab{}.
\newblock \showarticletitle{A Two-Stage Convolutional Neural Network for Hand Gesture Recognition}. In \bibinfo{booktitle}{\emph{Proceedings of the 6th International Conference on Advance Computing and Intelligent Engineering}}, \bibfield{editor}{\bibinfo{person}{Bibudhendu Pati}, \bibinfo{person}{Chhabi~Rani Panigrahi}, \bibinfo{person}{Prasant Mohapatra}, {and} \bibinfo{person}{Kuan-Ching Li}} (Eds.). \bibinfo{publisher}{Springer Nature Singapore}, \bibinfo{address}{Singapore}, \bibinfo{pages}{383--392}.
\newblock
\showISBNx{978-981-19-2225-1}


\bibitem[\protect\citeauthoryear{Maraqa and Abu-Zaiter}{Maraqa and Abu-Zaiter}{2008}]%
        {maraqa2008recognition}
\bibfield{author}{\bibinfo{person}{Manar Maraqa} {and} \bibinfo{person}{Raed Abu-Zaiter}.} \bibinfo{year}{2008}\natexlab{}.
\newblock \showarticletitle{Recognition of Arabic Sign Language (ArSL) using recurrent neural networks}. In \bibinfo{booktitle}{\emph{2008 First International Conference on the Applications of Digital Information and Web Technologies (ICADIWT)}}. IEEE, \bibinfo{pages}{478--481}.
\newblock


\bibitem[\protect\citeauthoryear{Maung}{Maung}{2009}]%
        {maung2009real}
\bibfield{author}{\bibinfo{person}{Tin Hninn~Hninn Maung}.} \bibinfo{year}{2009}\natexlab{}.
\newblock \showarticletitle{Real-time hand tracking and gesture recognition system using neural networks}.
\newblock \bibinfo{journal}{\emph{International Journal of Computer and Information Engineering}} \bibinfo{volume}{3}, \bibinfo{number}{2} (\bibinfo{year}{2009}), \bibinfo{pages}{315--319}.
\newblock


\bibitem[\protect\citeauthoryear{Min, Yoon, Soh, Yang, and Ejima}{Min et~al\mbox{.}}{1997}]%
        {min1997hand}
\bibfield{author}{\bibinfo{person}{Byung-Woo Min}, \bibinfo{person}{Ho-Sub Yoon}, \bibinfo{person}{Jung Soh}, \bibinfo{person}{Yun-Mo Yang}, {and} \bibinfo{person}{Toshiaki Ejima}.} \bibinfo{year}{1997}\natexlab{}.
\newblock \showarticletitle{Hand gesture recognition using hidden Markov models}. In \bibinfo{booktitle}{\emph{1997 IEEE International Conference on Systems, Man, and Cybernetics. Computational Cybernetics and Simulation}}, Vol.~\bibinfo{volume}{5}. IEEE, \bibinfo{pages}{4232--4235}.
\newblock


\bibitem[\protect\citeauthoryear{Molchanov, Gupta, Kim, and Kautz}{Molchanov et~al\mbox{.}}{2015a}]%
        {molchanov2015hand}
\bibfield{author}{\bibinfo{person}{Pavlo Molchanov}, \bibinfo{person}{Shalini Gupta}, \bibinfo{person}{Kihwan Kim}, {and} \bibinfo{person}{Jan Kautz}.} \bibinfo{year}{2015}\natexlab{a}.
\newblock \showarticletitle{Hand gesture recognition with 3D convolutional neural networks}. In \bibinfo{booktitle}{\emph{Proceedings of the International Conference on Computer Vision and Pattern Recognition (CVPR) Workshops}}. IEEE, \bibinfo{pages}{1--7}.
\newblock


\bibitem[\protect\citeauthoryear{Molchanov, Gupta, Kim, and Pulli}{Molchanov et~al\mbox{.}}{2015b}]%
        {molchanov2015multi}
\bibfield{author}{\bibinfo{person}{Pavlo Molchanov}, \bibinfo{person}{Shalini Gupta}, \bibinfo{person}{Kihwan Kim}, {and} \bibinfo{person}{Kari Pulli}.} \bibinfo{year}{2015}\natexlab{b}.
\newblock \showarticletitle{Multi-sensor system for driver's hand-gesture recognition}. In \bibinfo{booktitle}{\emph{Proceedings of the 11th International Conference on Automatic Face and Gesture Recognition}}. IEEE, \bibinfo{pages}{1--8}.
\newblock


\bibitem[\protect\citeauthoryear{Moniri and M\"{u}ller}{Moniri and M\"{u}ller}{2012}]%
        {Moniri2012a}
\bibfield{author}{\bibinfo{person}{Mohammad~Mehdi Moniri} {and} \bibinfo{person}{Christian M\"{u}ller}.} \bibinfo{year}{2012}\natexlab{}.
\newblock \showarticletitle{Multimodal reference resolution for mobile spatial interaction in urban environments}. In \bibinfo{booktitle}{\emph{Proceedings of the 4th International Conference on Automotive User Interfaces and Interactive Vehicular Applications}}. \bibinfo{publisher}{ACM}, \bibinfo{pages}{241–248}.
\newblock
\showISBNx{9781450317511}


\bibitem[\protect\citeauthoryear{Nagi, Ducatelle, Di~Caro, Cire{\c{s}}an, Meier, Giusti, Nagi, Schmidhuber, and Gambardella}{Nagi et~al\mbox{.}}{2011}]%
        {nagi2011max}
\bibfield{author}{\bibinfo{person}{Jawad Nagi}, \bibinfo{person}{Frederick Ducatelle}, \bibinfo{person}{Gianni~A Di~Caro}, \bibinfo{person}{Dan Cire{\c{s}}an}, \bibinfo{person}{Ueli Meier}, \bibinfo{person}{Alessandro Giusti}, \bibinfo{person}{Farrukh Nagi}, \bibinfo{person}{J{\"u}rgen Schmidhuber}, {and} \bibinfo{person}{Luca~Maria Gambardella}.} \bibinfo{year}{2011}\natexlab{}.
\newblock \showarticletitle{Max-pooling convolutional neural networks for vision-based hand gesture recognition}. In \bibinfo{booktitle}{\emph{2011 IEEE international conference on signal and image processing applications (ICSIPA)}}. IEEE, \bibinfo{pages}{342--347}.
\newblock


\bibitem[\protect\citeauthoryear{Ne{\ss}elrath, Moniri, and Feld}{Ne{\ss}elrath et~al\mbox{.}}{2016}]%
        {feld2016combine}
\bibfield{author}{\bibinfo{person}{Robert Ne{\ss}elrath}, \bibinfo{person}{Mohammad~Mehdi Moniri}, {and} \bibinfo{person}{Michael Feld}.} \bibinfo{year}{2016}\natexlab{}.
\newblock \showarticletitle{{Combining speech, gaze, and micro-gestures for the multimodal control of in-car functions}}. In \bibinfo{booktitle}{\emph{Proceedings of the 12th International Conference on Intelligent Environments}}. \bibinfo{publisher}{IEEE}, \bibinfo{pages}{190--193}.
\newblock


\bibitem[\protect\citeauthoryear{Neverova, Wolf, Paci, Sommavilla, Taylor, and Nebout}{Neverova et~al\mbox{.}}{2013}]%
        {neverova2013multi}
\bibfield{author}{\bibinfo{person}{Natalia Neverova}, \bibinfo{person}{Christian Wolf}, \bibinfo{person}{Giulio Paci}, \bibinfo{person}{Giacomo Sommavilla}, \bibinfo{person}{Graham Taylor}, {and} \bibinfo{person}{Florian Nebout}.} \bibinfo{year}{2013}\natexlab{}.
\newblock \showarticletitle{A multi-scale approach to gesture detection and recognition}. In \bibinfo{booktitle}{\emph{Proceedings of the IEEE International Conference on Computer Vision Workshops}}. \bibinfo{pages}{484--491}.
\newblock


\bibitem[\protect\citeauthoryear{Nickel, Scemann, and Stiefelhagen}{Nickel et~al\mbox{.}}{2004}]%
        {nickel20043d}
\bibfield{author}{\bibinfo{person}{Kai Nickel}, \bibinfo{person}{Edgar Scemann}, {and} \bibinfo{person}{Rainer Stiefelhagen}.} \bibinfo{year}{2004}\natexlab{}.
\newblock \showarticletitle{{3D-tracking of head and hands for pointing gesture recognition in a human-robot interaction scenario}}. In \bibinfo{booktitle}{\emph{Proceedings of the 6th International Conference on Automatic Face and Gesture Recognition}}. \bibinfo{publisher}{IEEE}, \bibinfo{pages}{565--570}.
\newblock


\bibitem[\protect\citeauthoryear{Nowak, Lukowicz, and Horodecki}{Nowak et~al\mbox{.}}{2018}]%
        {nowak2018assessing}
\bibfield{author}{\bibinfo{person}{Andrzej Nowak}, \bibinfo{person}{Paul Lukowicz}, {and} \bibinfo{person}{Pawel Horodecki}.} \bibinfo{year}{2018}\natexlab{}.
\newblock \showarticletitle{Assessing artificial intelligence for humanity: Will AI be the our biggest ever advance? Or the biggest threat [Opinion]}.
\newblock \bibinfo{journal}{\emph{IEEE Technology and Society Magazine}} \bibinfo{volume}{37}, \bibinfo{number}{4} (\bibinfo{year}{2018}), \bibinfo{pages}{26--34}.
\newblock


\bibitem[\protect\citeauthoryear{Ohn-Bar and Trivedi}{Ohn-Bar and Trivedi}{2014}]%
        {ohn2014hand}
\bibfield{author}{\bibinfo{person}{Eshed Ohn-Bar} {and} \bibinfo{person}{Mohan~Manubhai Trivedi}.} \bibinfo{year}{2014}\natexlab{}.
\newblock \showarticletitle{Hand gesture recognition in real time for automotive interfaces: A multimodal vision-based approach and evaluations}.
\newblock \bibinfo{journal}{\emph{IEEE Transactions on Intelligent Transportation Systems}} \bibinfo{volume}{15}, \bibinfo{number}{6} (\bibinfo{year}{2014}), \bibinfo{pages}{2368--2377}.
\newblock


\bibitem[\protect\citeauthoryear{Phyo, Fukuda, Lam, Kobayashi, and Kuno}{Phyo et~al\mbox{.}}{2019}]%
        {Phyo2019HRIGestures}
\bibfield{author}{\bibinfo{person}{Aye~Su Phyo}, \bibinfo{person}{Hisato Fukuda}, \bibinfo{person}{Antony Lam}, \bibinfo{person}{Yoshinori Kobayashi}, {and} \bibinfo{person}{Yoshinori Kuno}.} \bibinfo{year}{2019}\natexlab{}.
\newblock \showarticletitle{A Human-Robot Interaction System Based on Calling Hand Gestures}. In \bibinfo{booktitle}{\emph{Intelligent Computing Methodologies}}, \bibfield{editor}{\bibinfo{person}{De-Shuang Huang}, \bibinfo{person}{Zhi-Kai Huang}, {and} \bibinfo{person}{Abir Hussain}} (Eds.). \bibinfo{publisher}{Springer International Publishing}, \bibinfo{address}{Cham}, \bibinfo{pages}{43--52}.
\newblock
\showISBNx{978-3-030-26766-7}


\bibitem[\protect\citeauthoryear{Pickering, Burnham, and Richardson}{Pickering et~al\mbox{.}}{2007}]%
        {pickering2007research}
\bibfield{author}{\bibinfo{person}{Carl~A Pickering}, \bibinfo{person}{Keith~J Burnham}, {and} \bibinfo{person}{Michael~J Richardson}.} \bibinfo{year}{2007}\natexlab{}.
\newblock \showarticletitle{A research study of hand gesture recognition technologies and applications for human vehicle interaction}. In \bibinfo{booktitle}{\emph{Proceedings of the 3rd Institution of Engineering and Technology conference on automotive electronics}}. IET, \bibinfo{pages}{1--15}.
\newblock


\bibitem[\protect\citeauthoryear{Pinto, Borges, Almeida, and Paula}{Pinto et~al\mbox{.}}{2019}]%
        {pinto2019static}
\bibfield{author}{\bibinfo{person}{Raimundo~F Pinto}, \bibinfo{person}{Carlos~DB Borges}, \bibinfo{person}{Ant{\^o}nio Almeida}, {and} \bibinfo{person}{I{\'a}lis~C Paula}.} \bibinfo{year}{2019}\natexlab{}.
\newblock \showarticletitle{Static hand gesture recognition based on convolutional neural networks}.
\newblock \bibinfo{journal}{\emph{Journal of Electrical and Computer Engineering}}  \bibinfo{volume}{2019} (\bibinfo{year}{2019}).
\newblock


\bibitem[\protect\citeauthoryear{Rempel, Camilleri, and Lee}{Rempel et~al\mbox{.}}{2014}]%
        {Rempel2014}
\bibfield{author}{\bibinfo{person}{David Rempel}, \bibinfo{person}{Matt~J. Camilleri}, {and} \bibinfo{person}{David~L. Lee}.} \bibinfo{year}{2014}\natexlab{}.
\newblock \showarticletitle{{The design of hand gestures for human-computer interaction: Lessons from sign language interpreters}}.
\newblock \bibinfo{journal}{\emph{International Journal of Human Computer Studies}} \bibinfo{volume}{72}, \bibinfo{number}{10-11} (\bibinfo{date}{10} \bibinfo{year}{2014}), \bibinfo{pages}{728--735}.
\newblock
\showISSN{10959300}


\bibitem[\protect\citeauthoryear{Ren and Zhang}{Ren and Zhang}{2009}]%
        {ren2009hand}
\bibfield{author}{\bibinfo{person}{Yu Ren} {and} \bibinfo{person}{Fengming Zhang}.} \bibinfo{year}{2009}\natexlab{}.
\newblock \showarticletitle{Hand gesture recognition based on MEB-SVM}. In \bibinfo{booktitle}{\emph{2009 International Conference on Embedded Software and Systems}}. IEEE, \bibinfo{pages}{344--349}.
\newblock


\bibitem[\protect\citeauthoryear{Rigoll, Kosmala, and Eickeler}{Rigoll et~al\mbox{.}}{1997}]%
        {rigoll1997high}
\bibfield{author}{\bibinfo{person}{Gerhard Rigoll}, \bibinfo{person}{Andreas Kosmala}, {and} \bibinfo{person}{Stefan Eickeler}.} \bibinfo{year}{1997}\natexlab{}.
\newblock \showarticletitle{High performance real-time gesture recognition using hidden markov models}. In \bibinfo{booktitle}{\emph{International Gesture Workshop}}. Springer, \bibinfo{pages}{69--80}.
\newblock


\bibitem[\protect\citeauthoryear{Roider and Gross}{Roider and Gross}{2018}]%
        {roider2018see}
\bibfield{author}{\bibinfo{person}{Florian Roider} {and} \bibinfo{person}{Tom Gross}.} \bibinfo{year}{2018}\natexlab{}.
\newblock \showarticletitle{I see your point: Integrating gaze to enhance pointing gesture accuracy while driving}. In \bibinfo{booktitle}{\emph{Proceedings of the 10th International Conference on Automotive User Interfaces and Interactive Vehicular Applications}}. \bibinfo{publisher}{ACM}, \bibinfo{pages}{351–358}.
\newblock
\showISBNx{9781450359467}


\bibitem[\protect\citeauthoryear{Roider, R\"{u}melin, Pfleging, and Gross}{Roider et~al\mbox{.}}{2017}]%
        {Roider2017}
\bibfield{author}{\bibinfo{person}{Florian Roider}, \bibinfo{person}{Sonja R\"{u}melin}, \bibinfo{person}{Bastian Pfleging}, {and} \bibinfo{person}{Tom Gross}.} \bibinfo{year}{2017}\natexlab{}.
\newblock \showarticletitle{The effects of situational demands on gaze, speech and gesture input in the vehicle}. In \bibinfo{booktitle}{\emph{Proceedings of the 9th International Conference on Automotive User Interfaces and Interactive Vehicular Applications}}. \bibinfo{publisher}{ACM}, \bibinfo{pages}{94–102}.
\newblock
\showISBNx{9781450351508}


\bibitem[\protect\citeauthoryear{R{\"{u}}melin, Marouane, and Butz}{R{\"{u}}melin et~al\mbox{.}}{2013}]%
        {rumelin2013free}
\bibfield{author}{\bibinfo{person}{Sonja R{\"{u}}melin}, \bibinfo{person}{Chadly Marouane}, {and} \bibinfo{person}{Andreas Butz}.} \bibinfo{year}{2013}\natexlab{}.
\newblock \showarticletitle{{Free-hand pointing for identification and interaction with distant objects}}. In \bibinfo{booktitle}{\emph{Proceedings of the 5th International Conference on Automotive User Interfaces and Interactive Vehicular Applications}}. \bibinfo{publisher}{ACM}, \bibinfo{pages}{40--47}.
\newblock


\bibitem[\protect\citeauthoryear{Sainath, Vinyals, Senior, and Sak}{Sainath et~al\mbox{.}}{2015}]%
        {sainath2015convolutional}
\bibfield{author}{\bibinfo{person}{Tara~N Sainath}, \bibinfo{person}{Oriol Vinyals}, \bibinfo{person}{Andrew Senior}, {and} \bibinfo{person}{Ha{\c{s}}im Sak}.} \bibinfo{year}{2015}\natexlab{}.
\newblock \showarticletitle{Convolutional, long short-term memory, fully connected deep neural networks}. In \bibinfo{booktitle}{\emph{2015 IEEE international conference on acoustics, speech and signal processing (ICASSP)}}. IEEE, \bibinfo{pages}{4580--4584}.
\newblock


\bibitem[\protect\citeauthoryear{Shneiderman}{Shneiderman}{2020}]%
        {shneiderman2020human}
\bibfield{author}{\bibinfo{person}{Ben Shneiderman}.} \bibinfo{year}{2020}\natexlab{}.
\newblock \showarticletitle{Human-centered artificial intelligence: Reliable, safe \& trustworthy}.
\newblock \bibinfo{journal}{\emph{International Journal of Human--Computer Interaction}} \bibinfo{volume}{36}, \bibinfo{number}{6} (\bibinfo{year}{2020}), \bibinfo{pages}{495--504}.
\newblock


\bibitem[\protect\citeauthoryear{Singh}{Singh}{2015}]%
        {Singh2015}
\bibfield{author}{\bibinfo{person}{Dushyant~Kumar Singh}.} \bibinfo{year}{2015}\natexlab{}.
\newblock \showarticletitle{Recognizing hand gestures for human computer interaction}. In \bibinfo{booktitle}{\emph{Proceedings of the International Conference on Communications and Signal Processing}}. \bibinfo{publisher}{IEEE}, \bibinfo{pages}{379--382}.
\newblock


\bibitem[\protect\citeauthoryear{Stergiopoulou and Papamarkos}{Stergiopoulou and Papamarkos}{2009}]%
        {stergiopoulou2009hand}
\bibfield{author}{\bibinfo{person}{Ekaterini Stergiopoulou} {and} \bibinfo{person}{Nikos Papamarkos}.} \bibinfo{year}{2009}\natexlab{}.
\newblock \showarticletitle{Hand gesture recognition using a neural network shape fitting technique}.
\newblock \bibinfo{journal}{\emph{Engineering Applications of Artificial Intelligence}} \bibinfo{volume}{22}, \bibinfo{number}{8} (\bibinfo{year}{2009}), \bibinfo{pages}{1141--1158}.
\newblock


\bibitem[\protect\citeauthoryear{Stiegemeier, Bringeland, Kraus, and Baumann}{Stiegemeier et~al\mbox{.}}{2022}]%
        {Stiegemeier2022}
\bibfield{author}{\bibinfo{person}{Dina Stiegemeier}, \bibinfo{person}{Sabrina Bringeland}, \bibinfo{person}{Johannes Kraus}, {and} \bibinfo{person}{Martin Baumann}.} \bibinfo{year}{2022}\natexlab{}.
\newblock \showarticletitle{User Experience of In-Vehicle Gesture Interaction: Exploring the Effect of Autonomy and Competence in a Mock-Up Experiment}. In \bibinfo{booktitle}{\emph{Proceedings of the 14th International Conference on Automotive User Interfaces and Interactive Vehicular Applications}} (Seoul, Republic of Korea) \emph{(\bibinfo{series}{AutomotiveUI '22})}. \bibinfo{publisher}{Association for Computing Machinery}, \bibinfo{address}{New York, NY, USA}, \bibinfo{pages}{285–296}.
\newblock
\showISBNx{9781450394154}
\urldef\tempurl%
\url{https://doi.org/10.1145/3543174.3546847}
\showDOI{\tempurl}


\bibitem[\protect\citeauthoryear{Szegedy, Liu, Jia, Sermanet, Reed, Anguelov, Erhan, Vanhoucke, and Rabinovich}{Szegedy et~al\mbox{.}}{2015}]%
        {szegedy2015going}
\bibfield{author}{\bibinfo{person}{Christian Szegedy}, \bibinfo{person}{Wei Liu}, \bibinfo{person}{Yangqing Jia}, \bibinfo{person}{Pierre Sermanet}, \bibinfo{person}{Scott Reed}, \bibinfo{person}{Dragomir Anguelov}, \bibinfo{person}{Dumitru Erhan}, \bibinfo{person}{Vincent Vanhoucke}, {and} \bibinfo{person}{Andrew Rabinovich}.} \bibinfo{year}{2015}\natexlab{}.
\newblock \showarticletitle{Going deeper with convolutions}. In \bibinfo{booktitle}{\emph{Proceedings of the IEEE conference on computer vision and pattern recognition}}. \bibinfo{pages}{1--9}.
\newblock


\bibitem[\protect\citeauthoryear{Takahashi and Kishino}{Takahashi and Kishino}{1991}]%
        {Takahashi1991hand}
\bibfield{author}{\bibinfo{person}{Tomoichi Takahashi} {and} \bibinfo{person}{Fumio Kishino}.} \bibinfo{year}{1991}\natexlab{}.
\newblock \showarticletitle{Hand Gesture Coding Based on Experiments Using a Hand Gesture Interface Device}.
\newblock \bibinfo{journal}{\emph{SIGCHI Bull.}} \bibinfo{volume}{23}, \bibinfo{number}{2} (\bibinfo{date}{mar} \bibinfo{year}{1991}), \bibinfo{pages}{67–74}.
\newblock
\showISSN{0736-6906}
\urldef\tempurl%
\url{https://doi.org/10.1145/122488.122499}
\showDOI{\tempurl}


\bibitem[\protect\citeauthoryear{Takemura, Inui, and Fukui}{Takemura et~al\mbox{.}}{2018}]%
        {takemura2018neural}
\bibfield{author}{\bibinfo{person}{Naohiro Takemura}, \bibinfo{person}{Toshio Inui}, {and} \bibinfo{person}{Takao Fukui}.} \bibinfo{year}{2018}\natexlab{}.
\newblock \showarticletitle{A neural network model for development of reaching and pointing based on the interaction of forward and inverse transformations}.
\newblock \bibinfo{journal}{\emph{Developmental science}} \bibinfo{volume}{21}, \bibinfo{number}{3} (\bibinfo{year}{2018}).
\newblock
\newblock
\shownote{10 pages.}


\bibitem[\protect\citeauthoryear{Tsironi, Barros, and Wermter}{Tsironi et~al\mbox{.}}{2016}]%
        {tsironi2016gesture}
\bibfield{author}{\bibinfo{person}{Eleni Tsironi}, \bibinfo{person}{Pablo~VA Barros}, {and} \bibinfo{person}{Stefan Wermter}.} \bibinfo{year}{2016}\natexlab{}.
\newblock \showarticletitle{Gesture Recognition with a Convolutional Long Short-Term Memory Recurrent Neural Network.}. In \bibinfo{booktitle}{\emph{ESANN}}.
\newblock


\bibitem[\protect\citeauthoryear{Xu}{Xu}{2019}]%
        {xu2019toward}
\bibfield{author}{\bibinfo{person}{Wei Xu}.} \bibinfo{year}{2019}\natexlab{}.
\newblock \showarticletitle{Toward human-centered AI: a perspective from human-computer interaction}.
\newblock \bibinfo{journal}{\emph{interactions}} \bibinfo{volume}{26}, \bibinfo{number}{4} (\bibinfo{year}{2019}), \bibinfo{pages}{42--46}.
\newblock


\bibitem[\protect\citeauthoryear{Ye, Yang, and Xue}{Ye et~al\mbox{.}}{2018}]%
        {Ye2019}
\bibfield{author}{\bibinfo{person}{Qi Ye}, \bibinfo{person}{Lanqing Yang}, {and} \bibinfo{person}{Guangtao Xue}.} \bibinfo{year}{2018}\natexlab{}.
\newblock \showarticletitle{Hand-free gesture recognition for vehicle infotainment system control}. In \bibinfo{booktitle}{\emph{Proceedings of the IEEE Vehicular Networking Conference}}. \bibinfo{publisher}{IEEE}, \bibinfo{pages}{1--2}.
\newblock


\bibitem[\protect\citeauthoryear{Zhao, Wang, Liu, and Liu}{Zhao et~al\mbox{.}}{2019}]%
        {Zhao2019implGestureInfotain}
\bibfield{author}{\bibinfo{person}{Dan Zhao}, \bibinfo{person}{Cong Wang}, \bibinfo{person}{Yue Liu}, {and} \bibinfo{person}{Tong Liu}.} \bibinfo{year}{2019}\natexlab{}.
\newblock \showarticletitle{{Implementation and evaluation of touch and gesture interaction modalities for in-vehicle infotainment systems}}. In \bibinfo{booktitle}{\emph{Image and Graphics}}, \bibfield{editor}{\bibinfo{person}{Yao Zhao}, \bibinfo{person}{Nick Barnes}, \bibinfo{person}{Baoquan Chen}, \bibinfo{person}{Rüdiger Westermann}, \bibinfo{person}{Xiangwei Kong}, {and} \bibinfo{person}{Chunyu Lin}} (Eds.). \bibinfo{publisher}{Springer}, \bibinfo{pages}{384--394}.
\newblock
\showISBNx{978-3-030-34113-8}


\bibitem[\protect\citeauthoryear{Zimmerman, Lanier, Blanchard, Bryson, and Harvill}{Zimmerman et~al\mbox{.}}{1986}]%
        {Zimmerman1986hand}
\bibfield{author}{\bibinfo{person}{Thomas~G. Zimmerman}, \bibinfo{person}{Jaron Lanier}, \bibinfo{person}{Chuck Blanchard}, \bibinfo{person}{Steve Bryson}, {and} \bibinfo{person}{Young Harvill}.} \bibinfo{year}{1986}\natexlab{}.
\newblock \showarticletitle{A Hand Gesture Interface Device}. In \bibinfo{booktitle}{\emph{Proceedings of the SIGCHI/GI Conference on Human Factors in Computing Systems and Graphics Interface}} (Toronto, Ontario, Canada) \emph{(\bibinfo{series}{CHI '87})}. \bibinfo{publisher}{Association for Computing Machinery}, \bibinfo{address}{New York, NY, USA}, \bibinfo{pages}{189–192}.
\newblock
\showISBNx{0897912136}
\urldef\tempurl%
\url{https://doi.org/10.1145/29933.275628}
\showDOI{\tempurl}


\bibitem[\protect\citeauthoryear{Zobl, Geiger, Schuller, Lang, and Rigoll}{Zobl et~al\mbox{.}}{2003}]%
        {zobl2003real}
\bibfield{author}{\bibinfo{person}{Martin Zobl}, \bibinfo{person}{Michael Geiger}, \bibinfo{person}{Bj{\"o}rn Schuller}, \bibinfo{person}{Manfred Lang}, {and} \bibinfo{person}{Gerhard Rigoll}.} \bibinfo{year}{2003}\natexlab{}.
\newblock \showarticletitle{A real-time system for hand gesture controlled operation of in-car devices}. In \bibinfo{booktitle}{\emph{Proceedings of the International Conference on Multimedia and Expo (ICME)}}. IEEE.
\newblock
\newblock
\shownote{4 pages.}


\end{thebibliography}

\appendix

\end{document}